\documentclass[10pt,journal,compsoc]{IEEEtran}
\usepackage{multirow}
\usepackage{booktabs}
\usepackage{url}
\ifCLASSOPTIONcompsoc
  \usepackage[nocompress]{cite}
\else
  \usepackage{cite}
\fi
\ifCLASSINFOpdf
  \usepackage[pdftex]{graphicx}
\else
\fi
\usepackage{amsmath}
\usepackage{amsfonts}
\begin{document}
%
\title{Learning with Privileged Information via Adversarial Discriminative Modality Distillation}
%
%
%
%

\author{Nuno C. Garcia, Pietro Morerio, and Vittorio
Murino,~\IEEEmembership{Senior Member,~IEEE}

\thanks{\scriptsize N. C. Garcia, P. Morerio and V. Murino are with Pattern Analysis \&
Computer Vision (PAVIS), Istituto Italiano di Tecnologia (IIT), Genova, Italy.}
\thanks{\scriptsize N. C. Garcia is also with University of Genova, Italy.}
\thanks{\scriptsize V. Murino is also with Dept. of Computer Science, University of
Verona, Italy.}
\thanks{Primary email contact: \texttt{nuno.garcia@iit.it}.}
}

\IEEEtitleabstractindextext{%
\begin{abstract}
Heterogeneous data modalities can provide complementary cues for several tasks, usually leading to more robust
algorithms and better performance. However, while training data can be accurately collected to include a variety of sensory modalities,
it is often the case that not all of them are available in real life (testing) scenarios, where a model has to be deployed. This raises the
challenge of how to extract information from multimodal data in the training stage, in a form that can be exploited at test time,
considering limitations such as noisy or missing modalities. This paper presents a new approach in this direction for RGB-D vision
tasks, developed within the adversarial learning and privileged information frameworks. We consider the practical case of learning
representations from depth and RGB videos, while relying only on RGB data at test time. We propose a new approach to train a
hallucination network that learns to distill depth information via adversarial learning, resulting in a clean approach without several
losses to balance or hyperparameters. We report state-of-the-art results for object classification on the NYUD dataset, and video action
recognition on the largest multimodal dataset available for this task, the NTU RGB+D, as well as on the Northwestern-UCLA.
\end{abstract}

}

\maketitle

\IEEEdisplaynontitleabstractindextext

%
\IEEEpeerreviewmaketitle

\IEEEraisesectionheading{\section{Introduction}\label{sec:introduction}}

%
%
%
%



\IEEEPARstart{D}{epth} perception is the ability to reason about the 3D world, critical for the survival of many hunting predators and an important skill for humans to understand and interact with the surrounding environment. It develops very early in humans when babies start to crawl \cite{visualcliff}, and emerges from a variety of mechanisms that jointly contribute to the sense of relative and absolute position of objects, called depth cues. Besides binocular cues (\textit{e.g.} stereovision), humans use monocular cues that relate to \textit{a priori} visual assumptions derived from 2D single images through shadows, perspective, texture gradient, and other signals (\textit{e.g.} the assumption that objects look blurrier the further they are, or that if an object occludes another it must be closer, \textit{etc.}) \cite{watson2012depth}. As matter of fact, although humans underestimate object distance in a monocular vision setup \cite{Servos2000}, we are still able to perform most of our vision-related tasks with good efficiency even with one eye covered.

Similarly, depth perception is often of paramount importance for many computer vision tasks related to robotics, autonomous driving, scene understanding, to name a few. The emergence of cheap depth sensors and the need for big data led to big multimodal datasets containing RGB, depth, infrared, and skeleton sequences \cite{firman2016rgbd}, which in turn stimulated multimodal deep learning approaches. Traditional computer vision tasks like action recognition, object detection, or instance segmentation have been shown to benefit performance gains if the model considers other modalities, namely depth, instead of RGB only\cite{shahroudy2017deep, liu2017viewpoint, gupta2014learning, hazirbas2016fusenet}.

However, it is reasonable to expect that depth data is not going to be always available when a model is deployed in real scenarios, either due to the impossibility to collect depth data with enough quality (e.g., due to far-distance or reflectance issues) or to install depth sensors everywhere, sensor or communications failure, or other unpredictable events.

\begin{figure}[!t]
\centering
\includegraphics[width=\linewidth]{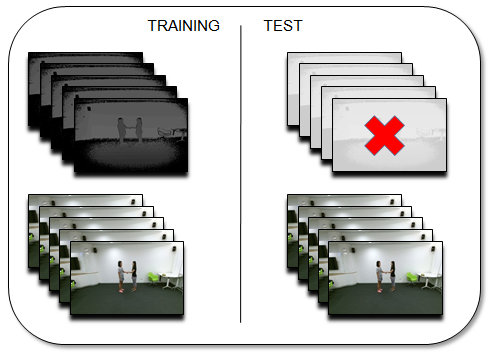}
\caption{What is the best way of using all data available at training time, considering a missing (or noisy) modality at test time?
}
\label{fig:intro}
\end{figure}

Considering this limitation, we would like to answer the following question (also depicted in Fig. 1): what is the best way of using all data available at training time, in order to learn robust representations, knowing that there are missing (or noisy) modalities at test time?
In other words, is there any added value in training a model by exploiting multimodal data, even if only one modality is available at test time?

Unsurprisingly, the simplest and most commonly adopted solution consists in training the model using only the modality in which it will be tested.
Nevertheless, a more interesting alternative is to exploit the potential of the available data and train the model using all modalities, being however aware of the fact that not all of them will be accessible at test time. This learning paradigm, \textit{i.e.}, when the model is trained using extra information, is generally known as \textit{learning with privileged information} \cite{vapnik2009new} or \textit{learning with side information} \cite{hoffman2016learning}. 

In this work, we propose an adversarial discriminative modality distillation (ADMD) strategy within a multimodal-stream framework that learns from different data modalities and can be deployed and tested on a subset of these. Particularly, our model learns from RGB \textit{and} depth video sequences and is tested on RGB only. 
Still, due to its general design it can also be used with whatever combination of other modalities as well. We evaluate its performance on the task of video action recognition and object classification. To this end, we introduce a new adversarial learning strategy to learn a hallucination network (Fig. \ref{fig:idea}), whose goal is to mimic the test time missing modality features, while preserving their discriminative power. The hallucination network uses RGB only as input and tries to recover useful depth features for the task at hand. Such network can be thought as a source of the aforementioned monocular depth cues, \textit{i.e.}, a source of depth cues from a single 2D RGB image. 

We would like to stress the fact that, in contrast to estimating real depth maps from RGB, we operate at feature level. Conceptually, it may seem that directly estimating depth maps from RGB is a more straightforward approach to deal with missing depth at test time. However, this is arguably a much more difficult task to accomplish compared to the primary task at hand, which is action/object recognition from RGB sequences. A more reasonable approach is to reduce the depth estimation problem from the pixel space to a low dimensional space, while continuing to profit to some extent of the discriminative benefits offered by the depth modality.

On the one hand, our work is inspired by previous works using hallucination networks in the context of learning with privileged information. This was primarily proposed in \cite{hoffman2016learning}, that presented an end-to-end single step training method to learn a hallucination network. This work was recently revisited in \cite{garcia2018modality} considering a multi-step learning paradigm using a loss inspired by the generalized distillation framework \cite{lopez2015unifying}. On the other hand, adversarial learning has been shown to be a powerful tool to model data distributions \cite{goodfellow2014generative, arora2018do}. Building upon these ideas, we propose a new approach to learn the hallucination network via a discriminative adversarial learning strategy. Our proposed method has several advantages: it is agnostic regarding the pair of modalities used, which greatly simplifies its extension beyond RGB and depth data; and it is able to deal with videos by design, by exploiting a form of temporal supervision as auxiliary information.
Furthermore, it dumps the need to balance the different losses used in the other methods \cite{hoffman2016learning} \cite{garcia2018modality}. 
Finally, thanks to the discriminator design, which includes an auxiliary classification task, our method is able to transfer the discriminative capability from a so-called \textit{teacher network}  \cite{lopez2015unifying} (depth network) to a \textit{student} (hallucination network), up to a full recovery of the teacher's accuracy.

To summarize, the main contributions of this paper are the following:
\begin{itemize}
\item We propose a new approach to learn a hallucination network within a multimodal-stream network architecture: it consists in an adversarial learning strategy that exploits multiple data modalities at training while using only one at test time. 
It proved to outperform its distance-based method counterparts \cite{hoffman2016learning, garcia2018modality}, and to augment 
its flexibility by being agnostic to components like distance metrics, data modalities, and size of the hallucinated feature vectors.
\item More technically, we propose a discriminator network which is time-aware, and jointly solves 1) the classical binary classification task (real/generated), and 2) an auxiliary task, which inherently endows the learned features with discriminative power.
\item We report results -- in the privileged information scenario -- on the NYUD \cite{NYUD} dataset for the task of object classification, and on the large-scale NTU RGB+D \cite{shahroudy2016ntu} and the Northwestern-UCLA \cite{wang2014cross} datasets for the task of action recognition.
\end{itemize}

The rest of the paper is organized as follows. Section \ref{sec:rela} relates this work to the literature in privileged information, multimodal deep learning, and adversarial learning.
Section \ref{sec:method} presents the details of the proposed architecture and the novel learning strategy. Section \ref{sec:exp} reports results on object recognition and video action recognition datasets, comparing them to the current state of the art, and investigating how the different parts of our approach contribute to the overall performance through an extensive ablation study. 
Finally, we draw conclusions and future research directions in Section \ref{sec:concl}.




\section{Related Work} \label{sec:rela}
Our work is at the intersection of four topics: adversarial learning \cite{goodfellow2014generative}, RGB-D vision, network distillation \cite{hinton2014distilling} and privileged information \cite{vapnik2009new}. As Lopez \emph{et al.} noted, privileged information and network distillation are instances of the same more inclusive theory, called generalized distillation \cite{lopez2015unifying}.

\subsection{Generalized Distillation} 
Within the generalized distillation framework, our model is both related to the privileged information theory \cite{vapnik2009new}, considering that the extra modality (depth, in this case) is only used at training time; and to the distillation framework, considering that our hallucination network is effectively learning by distilling the knowledge of a the previously learned "teacher" network, despite not using a distillation loss.

In this context, the closest works to our approach are \cite{hoffman2016learning} by Hoffman \textit{et al.} and \cite{garcia2018modality} by Garcia \textit{et al}.

The work of Hoffman \emph{et al.} \cite{hoffman2016learning} introduced a model to hallucinate depth features from RGB input for object detection task. While the idea of using a hallucination stream is similar to the one thereby presented, the mechanism used to learn it is different. In \cite{hoffman2016learning}, the authors use an Euclidean loss between the depth and hallucinated feature maps, that is part of the total loss along with more than ten classification and localization losses. This makes its effectiveness dependent on hyperparameter tuning to balance the different values, as the model is trained jointly in one step by optimizing the aforementioned composite loss.

In \cite{garcia2018modality}, Garcia \textit{et al.} built on this idea to propose a new staged training procedure that lead to learn a better teacher network, and a new loss inspired from the distillation framework. This loss is composed by the Euclidean distance between feature maps, the cross-entropy using the ground truth labels, and a cross-entropy using as targets the soft predictions from the teacher network (the depth stream, in this case). Moreover, the authors encouraged the learning by design, by using multiplier cross-stream connections \cite{feichtenhofer2017spatiotemporal}, and extended the method to video action recognition. Differently from \cite{garcia2018modality} and \cite{hoffman2016learning}, we propose an adversarial strategy to learn the hallucination stream, which alleviates the need for balancing losses or tuning hyperparameters.

An interesting work lying at the intersection of multimodal learning and learning with privileged information is ModDrop by Neverova \emph{et al.} \cite{neverova2016moddrop}. Here the authors propose a modality-based dropout strategy, where each input modality is \textit{entirely} dropped (actually zeroed) with some probability during training. The resulting model is proved to be more resilient to missing modalities at test time. We compare with ModDrop in the task of object classification.

Luo \emph{et al.} \cite{luo2017graph} addressed a similar problem, where the model is first trained on several modalities (RGB, depth, optical flow, and joints), but tested only in one. The authors propose a graph-based distillation method to distill information across all modalities at training time, allowing each modality to learn from all others. This approach achieves state-of-the-art results in action recognition and action detection tasks. 
Our work substantially differs from \cite{luo2017graph} since we benefit from a hallucination mechanism, consisting in an auxiliary hallucination network trained by leveraging a previously trained network. This mechanism allows the model to learn to emulate the presence of the missing modality at test time.

Another recent related work is \cite{zhang2018deep}, which proposes a distillation framework where there is no frozen teacher network, but all the networks work as an ensemble that learn in a collaboratively manner. Learning with privileged information for action recognition has also been explored for recurrent neural networks, \textit{e.g.} in \cite{shi2017learning} where the authors devise a method that uses skeleton joints as privileged information to learn a better action classifier that uses depth, even with scarce data.

\subsection{RGB-D vision}
Video action recognition and object detection have a long and rich field of literature, spanning from classification methods using handcrafted features, \textit{e.g.} \cite{dalal2005histograms, wang2013action,laptev2008learning,ye2013object,tang2012histogram,janoch2013category} to modern deep learning approaches, \textit{e.g.} \cite{karpathy2014large,tran2015learning,wang2017non,gupta2014learning,wang2015large}, using either RGB-only or together with depth data.
We point to some of the more relevant works in video action recognition and object recognition using RGB and depth, including state-of-the-art methods considering the NTU RGB+D and the NYU-Depth V2 datasets, as well as architectures related to our proposed model.

\subsubsection{Video action recognition} The two-stream model introduced by Simonyan and Zisserman \cite{simonyan2014two} is a landmark on video analysis, and since then has inspired a series of variants that achieved state-of-the-art performance on diverse datasets. This architecture is composed by a RGB and an optical flow stream, which are trained separately, and then fused at the prediction layer. Our model relates to this since the test-time predictions result from the average of the hallucination stream and the RGB stream logits.
In \cite{feichtenhofer2017spatiotemporal}, the authors propose a variation of the latter, which models spatiotemporal features by injecting the motion stream's signal into the residual unit of the appearance stream. They also employ 1D temporal convolutions along with 2D spatial convolutions, which we also adopt in this paper. Indeed, the combination of 2D spatial and 1D temporal convolutions has shown to learn better spatiotemporal features than 3D convolutions \cite{tran2018closer}. The current state of the art in video action recognition \cite{wang2018non} uses 3D temporal convolutions and a new building block dedicated to capture long range dependencies, using RGB data only.

Instead, in \cite{shahroudy2017deep} the authors explore the complementary properties of RGB and depth data, taking the NTU RGB+D dataset as testbed. They propose a deep autoencoder architecture and a structured sparsity learning machine, and achieve state-of-the-art results for action recognition. Liu \textit{et al.} \cite{liu2017viewpoint} also use RGB and depth to devise a method for viewpoint invariant action recognition. First, the method extracts dense trajectories from RGB data, which are then encoded in viewpoint invariant deep features. The RGB and depth features are then used as a dictionary for test time prediction. 

To the best of our knowledge, these are state-of-the-art approaches to exploit RGB+D for video action recognition, that report results on the NTU RGB+D dataset \cite{shahroudy2016ntu}, the largest video action recognition dataset to offer RGB and depth. It is important to note that we propose a fully convolutional model that exploits RGB and depth data at training time only, and uses exclusively RGB data as input at test time. This work goes in the direction of reducing the performance gap between privileged information and traditional approaches.

\subsubsection{Object recognition} Over the years, object recognition based on RGB and depth have been an insightful task to reason on the complementarity of these two modalities, and whether depth data should be handled differently, compared to RGB. An example of this is \cite{gupta2014learning}, in which the authors propose to encode depth images using a geocentric embedding that encodes height above ground and angle with gravity for each pixel in addition to the  horizontal disparity, showing that it works better than using raw depth. Differently, in \cite{wang2015large}, the authors focus on carefully designing a convolutional neural network including a multimodal layer to fuse RGB and depth.
Our work differs from these approaches since we focus on learning a model that has access to depth only at training time, which fundamentally changes the feature learning approach.

\subsection{Adversarial Learning}
In the seminal paper of Goodfellow \textit{et al.} \cite{goodfellow2014generative}, the authors propose a generative model that is trained by having two networks playing the so called minimax game. A generator network is trained to generate images from noise vectors, and a discriminator network is trained to classify the generated images as false, and images sampled from the dataset as true. As the game evolves, the generator becomes better and better at generating samples that look like the true images from the data distribution.

Many papers extended this approach in different directions, such as disentangling semantic concepts \cite{chen2016infogan}, network compression \cite{belagiannis2018adversarial} \cite{xu2018training} \cite{wang2018adversarial}, feature augmentation \cite{volpi2017adversarial}, image to image translation \cite{isola2017image}, and explored different losses \cite{arjovsky2017wasserstein} and other tricks to improve performance and stability \cite{salimans2016improved}\cite{radford2015unsupervised}. Our work relates to this body of work, as the hallucination network of our model tries to generate features from the missing modality feature space through adversarial learning. However, not only the context here is different, since adversarial learning is explored in the framework of privileged information, but also the task assigned to the discriminator is not the one typically used in adversarial learning, as detailed in Section \ref{sec:method}.

An important variant of the GAN framework are Conditional GANs (CGANs) \cite{mirza2014conditional}, that propose to concatenate the label of desired class to be generated, to the noise vector. This mechanism is related, yet different, to how our generator network is implicitly conditioned in this work. Our generator network input is a small volume of 5 RGB frames, and temporal convolutions are zero-padded to maintain the volume's size along the time dimension. The generator is thus implicitly aware of the temporal ordering, since features generated for the first and last frames will heavily be affected by the border effect of zero padding, which is performed several times (at each residual block). To solve this issue, we provide the temporal ordering conditioning label to the discriminator as well.

The CGAN model has been used in different domains, from image synthesis \cite{li2018fast} to domain adaptation \cite{volpi2017adversarial}.
Perhaps more similar to our work is the recent paper by Roheda \textit{et al.}\cite{roheda2018cross}, that also approaches the problem of missing modalities in the context of adversarial learning. They address the binary task of person detection using images, seismic, and acoustic sensors, where the latter two are absent at test time. A CGAN is conditioned on the available images and the generator maps a vector noise to representative information from the missing modalities, with an auxiliary L2 loss. In contrast to this work, our CGAN model learns a mapping directly from the test modality to the feature space of the missing modality, with no auxiliary loss. Besides, we propose a two-step training procedure in order to learn a better teacher network, and provide a stable target for the generator. Finally, we focus on the arguably more demanding tasks of video action recognition and object recognition.

\section{Learning to hallucinate depth features} \label{sec:method}

\begin{figure*}[!t]
\centering
\includegraphics[width=\textwidth]{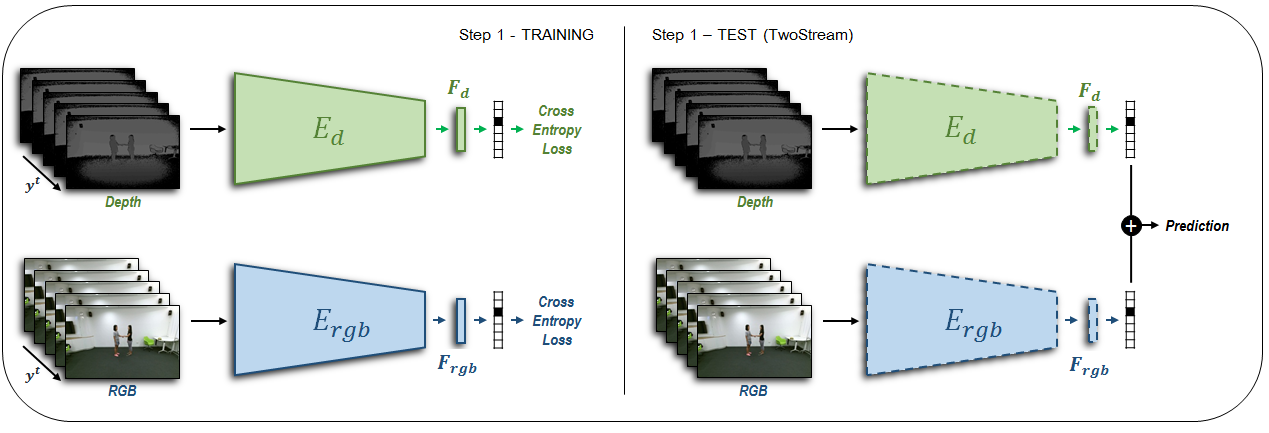}\\
\includegraphics[width=\textwidth]{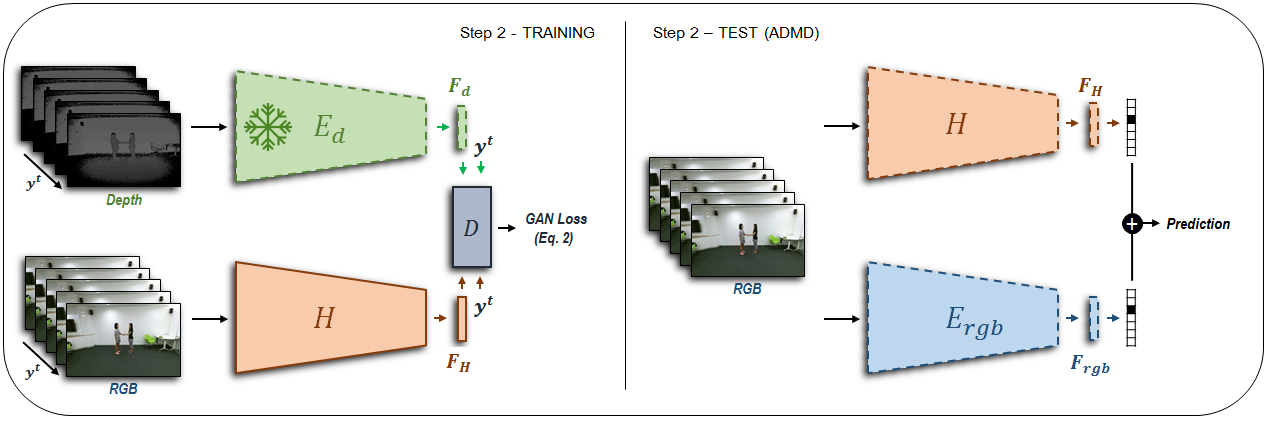}

\caption{Architecture and training steps (solid lines - module is \textit{trained}; dashed lines - module is \textit{frozen} ).  \textbf{Step 1:} Separate pretraining of RGB and Depth networks (Resnet-50 backbone with temporal convolutions). The bottleneck described in section \ref{subsec:archi} is highlighted as a separate component. At test time the raw predictions (logits) of the two separate streams are simply averaged. The complementary information carried by the two streams bring a significant boost in the recognition performance. \textbf{Step 2:} The depth stream is frozen. The hallucination stream $H$ is initialized with the depth stream's weights and adversarially trained against a discriminator. The discriminator is fed with the concatenation of the bottleneck feature vector and the temporal frame ordering label $y^t$, as detailed in Section \ref{subsec:training}. The discriminator also features an additional classification task, i.e. not only it is trained to discriminate between hallucinated and depth features, but also to assign samples to the correct class (Eq. \ref{eq:gan_loss}).
The hallucination stream thus learns monocular depth features from the depth stream while maintaining discriminative power. At test time, predictions from the RGB and the hallucination streams are fused.
}
\label{fig:idea}
\end{figure*}

Our goal is to train a hallucination network that, having as input RGB, is able to produce similar features to the ones produced by the depth network. The reasoning behind this idea is that on one hand depth and RGB provide complementary information for the task, but on the other hand RGB alone contains some cues for depth perception. 
Therefore, the goal of the hallucination network is to extract from RGB frames the complementary information that depth data would provide. It is important to emphasize that we are interested in recovering useful depth \textit{features}, in contrast to estimating real depth maps from RGB.

This is accomplished in a two-step training procedure, illustrated in Fig. \ref{fig:idea}, and described in the following. The \textit{first step} (\textit{Fig. \ref{fig:idea}, top}) consists in training the RGB and depth streams individually, with the respective input modality, as two standard, separate, supervised learning problems. The resulting ensemble, obtained by fusing the predictions of the two sub-networks (not fine-tuned), represents the full model (two-stream) that can be used when both modalities are available at test time. Its accuracy should be taken as an upper bound for the model we are proposing. In the \textit{second step} (\textit{Fig. \ref{fig:idea}, bottom}), we actually train the hallucination network by means of the proposed adversarial learning strategy. 
As the hallucination network is trained in the context of adversarial learning to generate depth features, it can be also interpreted as the generator network in the traditional GAN framework \cite{goodfellow2014generative}. However, strictly speaking, it is clearly to be considered as an encoder, which tries to extract monocular depth features from RGB input data. The test time setup of step 2 is again a two-stream model (not fine-tuned), composed by the RGB and hallucination networks, both having RGB data as input.

\subsection{Training procedure}\label{subsec:training}

Inspired by the generalized distillation paradigm, we follow a staged learning procedure, where the ``teacher'' net is trained first (Step 1) and separately from the ''student'' (Step 2). This is in contrast with \cite{hoffman2016learning}, where everything is learned end-to-end, but in line with \cite{garcia2018modality}, where separated learning steps proved to be more effective.

\textbf{Step 1.} The RGB and depth streams are trained separately, which is common practice in two-stream architectures. Both depth and appearance streams are trained by minimizing the cross-entropy loss, after being initialized with a pre-trained ImageNet model for all experiments as common practice \cite{garcia2018modality, hoffman2016learning, luo2017graph}. We test both streams individually and in a two-stream setup, where the final prediction results from the average of the two streams' logits. We found that fine-tuning the two-stream model does not increase performance consistently. This step can also be regarded as training the teacher network - depth stream - for the next step (see Fig. \ref{fig:idea}, top).

\textbf{Step 2.} The depth stream $E_{d}$, trained in the previous step, is now frozen, in order to provide a stable target for the hallucination network (generator) $H$, which plays the adversarial game with a discriminator $D$ (see Fig. \ref{fig:idea}, bottom). The primary task of the discriminator $D$ is to distinguish between the features $F_{H}$ generated by the hallucination network $H$ and $F_{d}$ generated by the depth network $E_{d}$. However, as already mentioned, the discriminator is also assigned an auxiliary discriminative task, as detailed in the following.

\medskip
The architecture of the networks $E_{d}$ and $H$ is a mix of 2D and 3D convolutions that process 
a set of frames, and output a feature vector for every frame $t$ of the input volume, \textit{i.e.} $F_{H}^{t}$ and $F_{d}^{t}$. This means that each frame have a corresponding feature vector, and these may vary even if sampled from the same video, depending on its dynamics and its position $t$ in the input volume. For example, the first frame (and feature vector) of a clip belonging to the action "shaking hand" might be very different from its the middle frame, but similar to the first frame of a clip belonging to the class "pushing other person". This increases the complexity for the generator, that have not only to generate features similar to $F_{d}$, but also to match the order in which they are generated. Namely, $F_{H}^{t}$ should be similar to $F_{d}^{t}$, for every frame $t$ of the input volume. We ease this issue by providing as input to $D$ the one-hot encoding vector of the relative index $t$, which we denote $y^t$, concatenated with the respective feature vector, which relates to the CGAN mechanism \cite{mirza2014conditional}.

In standard adversarial training, the discriminator $D$ would try to assign the binary label true/fake to the feature vector coming from the two different streams. However, we found that features $F_{H}$ generated with this mechanism, although being very well mixed and indistinguishable from $F_{d}$, were struggling to achieve good accuracy for the classification tasks, \textit{i.e.} were lacking discriminative power. For this reason we assign to the discriminator the auxiliary task of classifying feature vectors with their correct class.

\medskip
The adversarial learning problem is formalized as follows. 
Consider the RGB-D dataset $(X_{rgb}, X_{d}, Y)$  where $x_{rgb}^t, x_{d}^t \sim (X_{rgb}, X_{d})$ are time aligned RGB and depth frames, $y\sim Y$, is the $C$-dimensional one-hot encoding of the class label, and $C$ is the number of classes for the problem at hand.

Now, let the \textit{extended label vector} with $C+1$ components (classes):
\begin{equation}\label{eq_extendedlabel}
\hat{y} = 
\begin{cases} [zeros(C)\,||\,1], & \mbox{for } x_{rgb} \\ 
[y_i\,||\,0] & \mbox{for } x_d 
\end{cases}
\end{equation}
where $zeros(C)$ represents a vector of zeros of dimension C, and $||$ is the concatenation operator.
Using this label vector instead of the classical 0/1 (real/generated) binary label in the discriminator encourages feature representations $F_H$ learned by $H$ to encode not only depth (monocular) features, but also to be discriminative. This is possibly why the hallucination network often recovers the accuracy of the teacher and sometimes performs even better, as further discussed in the experimental section. In summary, we want $F_H$ features to be as discriminant as real ones: the adversarial procedure produces fake features which not only are classified as real by the discriminator, but are also assigned to the correct class.

Based on the above definitions, we define the following minimax game:
\begin{equation} \label{eq:gan_loss}
\begin{split}
\min_{\theta_D} \max_{\theta_{H}} \ell &= \mathbb{E}_{(x_i,y_i) \sim (X_{rgb},Y)} \mathcal{L}( D(H(x_i) \vert\vert y^t), \; \hat{y}_i ) \\
& + \mathbb{E}_{(x_i,y_i) \sim (X_d,Y)} \mathcal{L}(\vert D(E_d(x_i) \vert\vert y^t), \;\hat{y}_i) 
\end{split}
\end{equation}

where $\theta_H$ and $\theta_{D}$ indicate the parameters of the hallucination stream $H$ and of the discriminator $D$, $\vert\vert$ denotes a concatenation operation and $\mathcal{L}$ is the softmax cross-entropy function. Eq. \ref{eq:gan_loss} is optimized via the well known ''label flipping hack'' \cite{ganhacks}, which makes the loss function easier to minimize in practice.

\subsection{Architectural details}\label{subsec:archi}

All three networks (depth stream - $E_d$, RGB stream - $E_{rgb}$, and hallucination stream $H$) are modified Resnet-50 \cite{he2016deep} augmented with\textit{ temporal convolutions} and endowed with a final \textit{bottleneck layer}. The hallucination networks $H$ are initialized with the respective depth stream weights $E_d$, following the findings of \cite{hoffman2016learning} for object detection, and \cite{garcia2018modality} for action recognition. 

\begin{figure}[!b]
\centering
\includegraphics[width=\linewidth]{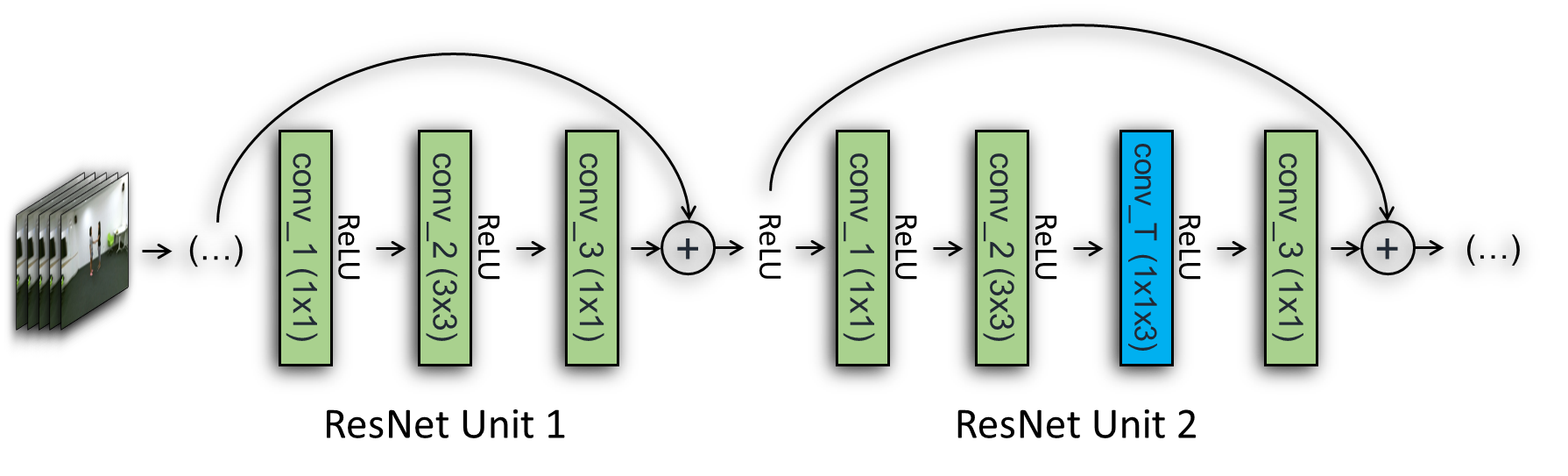}
\caption{Detail of the ResNet residual unit with temporal convolutions (blue block).}
\label{fig:temp}
\end{figure}

\textbf{Temporal convolutions.} 1D temporal convolutions are inserted in the second residual unit of each ResNet layer as illustrated in Fig. \ref{fig:temp}, following the recent work of Feichtenhofer \textit{et al.} \cite{feichtenhofer2017spatiotemporal}.
For layer $l$, the weights $W_{l} \in \mathbb{R}^{1 \times 1 \times 3 \times C_{l} \times C_{l}}$ are convolutional filters initialized as identity mappings at feature level, and centered in time, where $C_{l}$ is the number of channels in layer $l$. More in detail, all the $[1\times 1\times 3]$ temporal kernels contained in $W_{l}$  are initialized as $[0,1,0]$, \textit{i.e.} only the information of the central frame is used at the beginning. This progressively changes as training goes on. 
Very recently, in \cite{tran2018closer}, the authors explored various network configurations using temporal convolutions, comparing different combinations for the task of video classification. This work suggests that decoupling 3D convolutions into 2D (spatial) and 1D (temporal) filters is the best setup in action recognition tasks, producing best accuracies. The intuition for the latter setup is that factorizing spatial and temporal convolutions in two consecutive convolutional layers eases training  of the spatial and temporal tasks (also in line with \cite{sun2015human}).

\textbf{Bottleneck.} Generating, encoding, or aligning high dimensional feature vectors via adversarial training is often a difficult task, due to the inherent instability of the saddle point defined by the GAN minimax game. For this reason, \cite{volpi2017adversarial} proposes to align a lower dimensional vector, obtained by adding a \textit{bottleneck layer} to standard architectures. This usually does not affect performances of baseline models.

Indeed, the size of the last ResNet-50 layer (before the logits) is $[7,7,2048]$, or simply $[2048]$ after pooling. For this reason, we further modify the ResNet-50 by adding either i) an additional convolutional layer, whose weights $W_{b} \in \mathbb{R}^{7 \times 7 \times 2048 \times 128 }$, applied with no padding, reduce the dimensionality to $128$; or ii) a simple 128-dim fully connected layer after pooling. In Section \ref{subsub:bottleneck} we further explore the choice of the bottleneck.

\textbf{Input.} For the task of action recognition, the input to the encoder networks $E$ and $H$ is five 3-channel frames, uniformly sampled from each video sequence, which motivates temporal convolution. Instead, for the task of object classification (from single images), no temporal kernels are added to the architecture. 
We try  different encodings for the depth channel: for the task of action recognition they are encoded into color images using a jet colormap, as in \cite{eitel2015multimodal}; for the object recognition task, HHA encoding \cite{HHA} is already provided in the dataset considered.

\begin{figure}[!t]
\centering
\includegraphics[width=.8\linewidth]{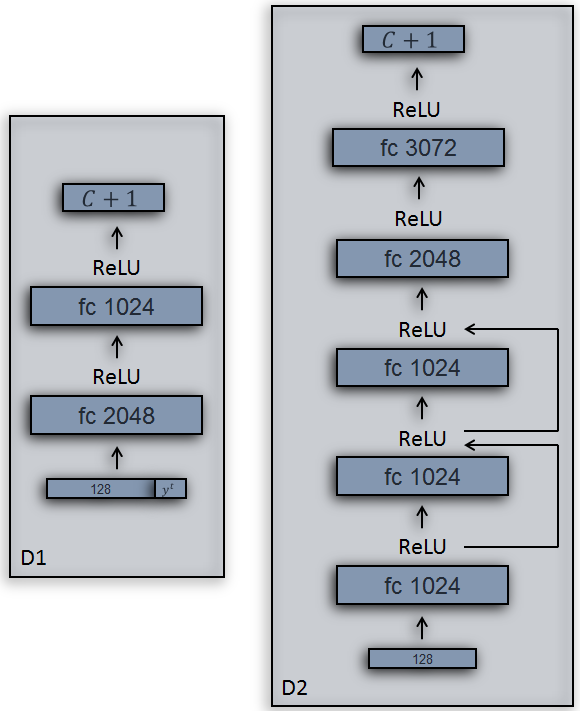}
\caption{Architectures for the discriminators used for the two different tasks. Left: D1 for object recognition. Right: D2 for action recognition.}
\label{fig:disc}
\end{figure}

\textbf{Discriminator.}
The discriminator used to play the adversarial game has different architectures depending on the task. These architectures follow the empirically validated common practices in the adversarial learning literature, more specifically to what is described in \cite{volpi2017adversarial}.
Its basic structure is that of a multilayer perceptron, stacking fully connected (fc) layers only, since it takes a vector as input (bottleneck features, possibly concatenated with temporal ordering for tasks involving time). 
For the task of action recognition, the structure is quite shallow, consisting in D1=[fc(2048), fc(1024), fc($C+1$)].  For the task of object classification the structure is instead more complex D2=[fc(1024), fc(1024), fc(1024), fc(2048), fc(3072), fc($C+1$)], with skip connections in the lower layers. Being the former discriminator quite deep, residual connections were inserted in order to allow gradient to flow through the underlying hallucination stream. Details of the architectures are sketched in Fig. \ref{fig:disc}.

\section{Experiments} \label{sec:exp}
\subsection{Datasets}
We evaluate the performance of our method on one object classification and two video action classification datasets. For both tasks the model is initialized with ImageNet pretrained weights. For the experiments on the smaller action recognition dataset NW-UCLA, we fine-tune the model starting from the RGB and depth streams trained on the larger NTU RGB+D dataset.

\textbf{NTU RGB+D \cite{shahroudy2016ntu}.} This is the largest public dataset for multimodal video action recognition. It is composed by 56,880 videos, available in four modalities: RGB videos, depth sequences, infrared frames, and 3D skeleton data of 25 joints (RGB and depth examples illustrated in Fig. \ref{fig:intro}). It was acquired with a Kinect v2 sensor in 80 different viewpoints, and includes 40 subjects performing 60 distinct actions.
We follow the two evaluation protocols originally proposed in \cite{shahroudy2016ntu}, which are cross-subject and cross-view. As in the original paper, we use about 5\% of the training data as validation set for both protocols. The masked depth maps are converted to a three channel map via a jet mapping, as in \cite{eitel2015multimodal}.

\textbf{Northwestern-UCLA \cite{wang2014cross}.}
This action recognition dataset provides RGB, depth and skeleton sequences for 1475 samples. It features 10 subjects performing 10 actions captured in 3 different views.

\begin{figure}[t!]
	\centering
	\begin{tabular}{@{}l@{}l@{}l@{}l@{}l}
	\includegraphics[width=0.18\linewidth]{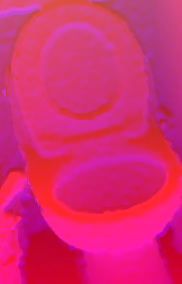}\; &
	\includegraphics[width=0.18\linewidth]{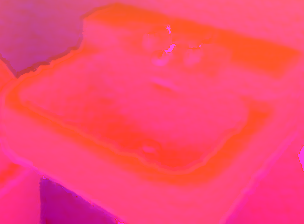}\;&
	\includegraphics[width=0.18\linewidth]{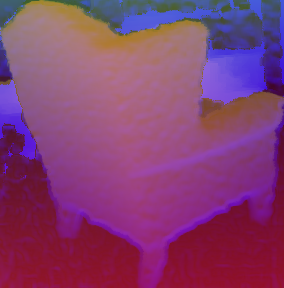}\;&
	\includegraphics[width=0.18\linewidth]{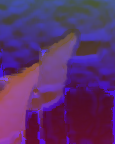}\;&
	\includegraphics[width=0.18\linewidth]{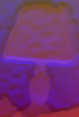}\\
	\includegraphics[width=0.18\linewidth]{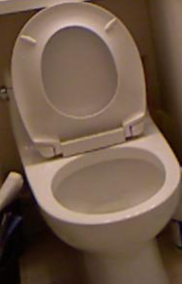}&
	\includegraphics[width=0.18\linewidth]{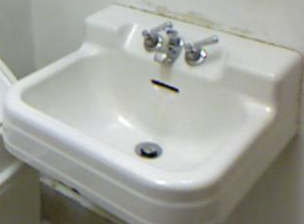}&
	\includegraphics[width=0.18\linewidth]{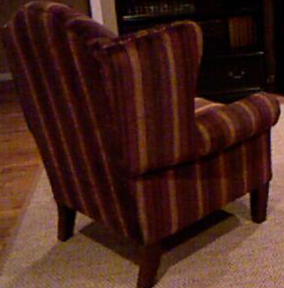}&
	\includegraphics[width=0.18\linewidth]{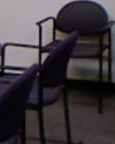}&
	\includegraphics[width=0.18\linewidth]{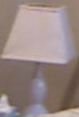}
	\end{tabular}

	\caption{Examples of RGB and depth frames from the NYUD (RGB-D) dataset.}
	\label{fig:nyud}
\end{figure}

\textbf{NYUD (RGB-D)} This dataset of objects (see examples in Fig. \ref{fig:nyud}) is gathered by cropping out tight bounding boxes around instances of 19 object classes present in the NYUD \cite{NYUD} dataset. It comprises 2,186 paired labeled training images and 2,401 test images (RGB-D). Depth images are HHA-encoded \cite{HHA}. This version of the dataset was proposed in \cite{hoffman2016learning} but also used in \cite{Tzeng2017adda, volpi2017adversarial, morerio2018minimalentropy} for the task of modality adaptation, in the framework of domain adaptation (train on one modality, adapt and test the model on the other modality). The task here is object classification, training on both modalities and testing on RGB only.

\subsection{Ablation Study}
The ablation study is performed on part of the NTU RGB+D dataset, designated as mini-NTU, which consists of random samples from the training set, 
considering approximately a third of the original dataset size. The test set is still the same as used in the other experiments and defined originally in \cite{shahroudy2016ntu}.

We study how the hallucination network performance is affected by (1) feeding different types of input to the discriminator, and (2) having the discriminator to perform different tasks.

\subsubsection{Bottleneck size}\label{subsub:bottleneck}
The discriminator receives as input the feature vector $F_H$ or $F_d$ from either the hallucination or the depth stream, respectively, along with the frame index label $y^t$. It is known that a too big feature vector may cause the GAN training to underperform \cite{volpi2017adversarial}, which we also observe in our experiments, reported in Table \ref{table_bott_sz}.

We first trained our depth network without bottleneck on the full NTU dataset, reaching 70.53\% accuracy. This network is then  used as target to learn the hallucination model. We observed that the hallucination model trained without bottleneck, \textit{i.e.}, the input to the discriminator is the 2048-dimensional feature vector, is far from recovering the performance of the target (reaching only 54.25\%), even if the training space is reduced to the NTU-mini dataset (60.95\%).

We then train a network with a 128-dimensional bottleneck (69.13\%), initialized with the previous depth stream, except for the bottleneck that is randomly initialized with the MSRA initialization \cite{he2015delving}. 
The hallucination model that learns using the bottleneck feature vector is able not only to recover, but to surpass the performance of the depth stream, reaching 72.14\% accuracy. We observed this behaviour in other experiments along the paper, and we comment that later in Section \ref{comp:obj}.

\begin{table}
\caption{Ablation Study - Bottleneck size. Hallucination network underperforming with $F_x \in \mathbb{R}^{2048}$.}
\label{table_bott_sz}
\centering
\footnotesize
\begin{tabular}{ccc}
\toprule
Network  & Dataset & $X$-Subject\\
\midrule
Depth stream, normal - (target) & NTU & 70.53\% \\
Hall. net, $F_x \in \mathbb{R}^{2048}$  & NTU & 54.25\% \\
Hall. net, $F_x \in \mathbb{R}^{2048}$ & NTU-mini & 60.95\% \\
\midrule
Depth stream, w/ bottleneck - (target) & NTU & 69.13\% \\
Hall. net, $F_x \in \mathbb{R}^{128}$ & NTU & 72.14\% \\
\bottomrule
\end{tabular}
\end{table}

\subsubsection{Bottleneck implementation}
In Table \ref{table_bott} we investigate different ways to decrease the size of $F_{x}$ from $\mathbb{R}^{2048}$ to $\mathbb{R}^{128}$, as suggested in \cite{volpi2017adversarial}. After the last feature map, which is of dimension 7*7*2048, we tested the three following ways:
\begin{itemize}
\item convolution of [128,7,7] to 1*1*128,
\item spatial convolution of [7,7] to 1*1*2048 followed by 1D convolution to 1*1*128, and
\item pooling layer to 1*1*2048 followed by 1D convolution to 1*1*128
\end{itemize}

Even though the depth stream is just trained on the NTU-mini (63.95\% for cross subject, and 62.70\% for cross view), the hallucination stream that implements the pool+conv bottleneck is able to recover almost completely (61.41\% for cross subject), or even surpass (63.15\% for cross view), the original depth stream performance. This was the architectural choice we used in the rest of the experiments.

\begin{table}[!t]
\caption{Ablation Study -  Investigating different bottleneck implementations. The Table reports Hallucination network performances on NTU-mini.}
\label{table_bott}
\centering
\begin{tabular}{ccc}
\toprule
Depth stream - versions & $X$-Subject & $X$-View\\
\midrule
Depth stream wo/ bottleneck & 63.95\% & 62.70\% \\
One conv & 55.64\% & 57.91\% \\
Spatial conv + 1D conv & 53.21\% & 52.58\% \\
pool + conv & 61.41\% & 63.15\% \\
\bottomrule
\end{tabular}
\end{table}


\subsubsection{Discriminator: inputs and tasks} 
In this section, we explore whether the task assigned to the discriminator influences the hallucination performance. As introduced in Section \ref{sec:method}, our hypothesis is that the generator has the difficult task of generating features that not only correspond to depth features, but also need to be temporally paired with these. We solve this by introducing the additional information of the frame index $y^t$, which specifies the desired alignment. Table \ref{table_disc} shows results regarding the (1) traditional binary task of a GAN generator having as input the feature bottleneck, (2) the $\hat{y}$ classification task having the same input as before, and (3) the proposed approach. The traditional binary task (1) converges to a perfect equilibrium, but the hallucination stream's accuracy is close to random chance, meaning that the learned features are not discriminant at all. The second approach (2) is able to learn discriminative features, but the addition of the frame order supervision $y^t$ (3) shows an increase in performance. It is reasonable that this mechanism produces maximized gains on more challenging and diverse datasets, as the full NTU dataset, or in fully 3d-convolutional architectures such as I3D \cite{carreira2017quo}, due to the higher dependence on temporal convolutions.

\begin{table}[!t]
\caption{Ablation Study - Investigating different inputs and tasks for the discriminator. The Table reports Hallucination network performances (NTU-mini).}
\label{table_disc}
\centering
\begin{tabular}{cccc}
\toprule
Input & Task & $X$-Subject\\
\midrule
Teacher network & \multirow{2}{*}{-} & \multirow{2}{*}{61.41\%}\\
(pool + conv, Table \ref{table_bott}) && \\
\midrule
F(x) & 0/1 classification & 1.81\% \\
F(x) & $\hat{y}$ classification  & 59.87\% \\
F(x) $\vert\vert y_t$ & $\hat{y}$ classification & 63.03\% \\
\bottomrule
\end{tabular}
\end{table}

\subsection{Action recognition performance and comparisons}
\begin{table*}[h!tb]
    \caption{Classification accuracies and comparisons with the state of the art for video action recognition. Performances referred to the several steps of our approach (ours) are highlighted in bold. $\times$ refers to comparisons with unsupervised learning methods.  $\bigtriangleup$ refers to supervised methods: here train and test modalities coincide. $\Box$ refers to privileged information methods: here training exploits RGB+D data, while test relies on RGB data only. The 4th column refers to cross-subject and the 5th to the cross-view evaluation protocols on the NTU dataset. The results reported on the other two datasets are for the cross-view protocol.}\label{table:sota1}
	\begin{center}
		\begin{tabular}{lllcccc}
			\# & Method & Test Mods. & NTU (p1) & NTU (p2) & NW-UCLA &\\
            \toprule
			1 & Luo \cite{luo2017unsupervised} & Depth & 66.2\% & - &-& \multirow{3}{*}{$\times$}\\
			2 & Luo \cite{luo2017unsupervised} & RGB & 56.0\% & - &-&\\
            3 & Rahmani \cite{rahmani2018learning} & RGB & - & - & 78.1\% & \\
			\midrule
			4 & HOG-2 \cite{ohn2013joint} & Depth & 32.4\% & 22.3\% &-& \multirow{15}{*}{$\bigtriangleup$}\\
            5 & Action Tube \cite{gkioxari2015finding} & RGB & - & - & 61.5\% &\\
            6 & Depth stream \cite{garcia2018modality} & Depth & 70.44\% & 75.16\% & 72.38\% &\\
            7 & \textbf{ADMD} - Depth stream & Depth & 70.53\% & 76.47\% & -  &\\
            8 & \textbf{ADMD} - Depth stream w/ bott. & Depth & 71.87\% & 75.32\%  & 71.09\% &\\
			9 & \cite{garcia2018modality} - RGB stream  & RGB & 66.52\% & 80.01\%  & 85.22\%&\\
			10 & \textbf{ADMD} - RGB stream  & RGB & 67.95\% & 80.01\% &  85.87\%&\\
			11 & Deep RNN \cite{shahroudy2016ntu} & Joints & 56.3\% & 64.1\% &-&\\
			12 & Deep LSTM \cite{shahroudy2016ntu} & Joints &   60.7\%  & 67.3\% &-&\\
			13 & Sharoudy \cite{shahroudy2016ntu} & Joints &   62.93\%  & 70.27\% &-&\\
			14 & Kim \cite{soo2017interpretable} & Joints & 74.3\% & 83.1\% &-&\\
			15 & Sharoudy \cite{shahroudy2017deep} & RGB+D &  74.86\%  & - &-&\\
			16 & Liu \cite{liu2017viewpoint} & RGB+D & 77.5\% & 84.5\% &-&\\
            17 & Rahmani \cite{rahmani2017learning} & Depth+Joints & 75.2 & 83.1 & - & \\
            18 & Two-stream, step 2 \cite{garcia2018modality} & RGB+D & 79.73\% & 81.43\% & 88.87\% &\\
            19 & \textbf{ADMD} - Two-stream (no finetune) & \textbf{RGB+D} & \textbf{77.74}\% & \textbf{85.49}\% & \textbf{89.93\%} &\\ 
			\midrule
			\midrule
            20 & Hoffman \emph{et al.} \cite{hoffman2016learning} & RGB & 64.64\% & -& 83.30\%& \multirow{7}{*}{$\Box$}\\
            21 & Luo \emph{et al.} \cite{luo2017graph} & RGB & 89.50\% & - & - &\\ 
			22 & Hallucination model, step 3 \cite{garcia2018modality} & RGB & 71.93\% & 74.10\% & 76.30\% &\\ 
            23 & Hallucination model, step 4 \cite{garcia2018modality} & RGB & 73.42\% & 77.21\% & 86.72\% &\\ 
            \cmidrule{1-6}
            24 & \textbf{ADMD} - Hall. stream alone& \textbf{RGB} & \textbf{67.57}\% & \textbf{71.80}\% & \textbf{83.94\%} &\\ 
            25 & \textbf{ADMD} - Hall. two-stream model & \textbf{RGB} & \textbf{73.11}\% & \textbf{81.50}\% & \textbf{91.64\%} &\\ 
            \bottomrule
		\end{tabular}
	\end{center}
\end{table*}

Table \ref{table:sota1} compares performances of different methods in the literature, across the two datasets for action recognition - two protocols for the NTU RGB+D and the NW-UCLA. The standard performance measure used for this task and datasets is classification accuracy, estimated according to the protocols, training and testing splits defined in the respective works.
The first part of the table (indicated by $\times$ symbol) refers to unsupervised methods, which achieve surprisingly high results even without relying on labels in learning representations.

The second part refers to supervised methods (indicated by $\bigtriangleup$), divided according to the modalities used for training and testing. Here, we report the performance of the separate RGB and depth (with and without bottleneck) streams trained in step 1 (rows \#7 and \#8). The small increase in performance is probably due to the extra training steps with small learning rate, after initialized with the bottleneck version trained on the mini-NTU (used for the ablation study). Importantly, the depth stream with bottleneck represents the teacher network used for the hallucination learning. We expect our final model to perform better than the one trained on RGB only, whose accuracy constitutes a lower bound for the usefulness of our hallucination model. The values reported for our step 1 models for the NW-UCLA dataset, \textit{i.e.} the RGB and depth streams, refer to the fine-tuning of our NTU model. In contrast with \cite{garcia2018modality}, and for clearer analysis, the two-stream setup is always not finetuned. Its accuracy represents an upper bound for the final model, which will not rely on depth data at test time. We have experimented training using pre-trained ImageNet weights instead of the NTU, but it led to lower accuracy.

The last part of the table (indicated by $\Box$) reports the performance of methods in the privileged information framework, thus directly comparable to ours. 
The performance values that refer to the Hoffman \textit{et al.} method \cite{hoffman2016learning} (row \#20 of Table \ref{table:sota1}) are taken from the implementation and experiments in \cite{garcia2018modality}. Row \#21 refers to the method by Luo and colleagues \cite{luo2017graph}, that uses 6 modalities at training time (RGB, depth, optical flow, and three different encoding methods for skeleton data), and RGB only at test time. Step 3 and 4 of \cite{garcia2018modality} (row \#22 and \#23) refer to the two-stream model after the hallucination learning, and its fine-tuning, respectively. We note that, for simplicity, the results of ADMD Two-Stream models are merely the outcome of the average of the two streams' logits, and are not subject to any fine-tuning, which means that they are directly comparable with row \#22. 
In addition, results of row \#24 correspond to the hallucination stream only.

We note that the hallucination stream (row \#24) manages to recover and surpass the depth teacher stream (row \#8) for the NW-UCLA dataset (83.94\% compared to 71.09\%), while for the NTU p1 (67.57\%) and p2 (71.80\%) protocols is around 4\% below the respective teacher (71.87\% and 75.32\%). Nevertheless, when combined with the RGB stream, it performs better (NTU p2 - 81.50\%) or comparable (NTU p1 - 73.11\%) to the fine-tuned model presented in \cite{garcia2018modality}. Since the RGB stream is performing equally well in this work and in \cite{garcia2018modality}, we can conclude that the gains in performance are due to better hallucination features.

\subsection{Object recognition performance and comparisons}
\label{comp:obj}
Table \ref{table_objrec} illustrates the main results obtained for NYUD dataset for the object recognition task.

As opposed to action recognition, depth information is often noisy here (cfr. Fig. \ref{fig:nyud} - chair and lamp), probably due to the small resolution of the bounding box crops. Depth alone is in fact performing worse than RGB alone (more than 10\% gap). Still, the amount of \textit{complementary information} carried by the two modalities is able, when fused in the two-stream model, to boost recognition accuracy by more than 5 percentage points, despite the poor depth performance (RGB$\rightarrow$52.90\%, Depth$\rightarrow$40.19\% $\Rightarrow$ two-stream$\rightarrow$57.39\%). 

\begin{table}[hbt]
\caption{Object Recognition}
\label{table_objrec}
\centering
\setlength\tabcolsep{4pt}
\begin{tabular}{lccc}
Method & Trained on & Tested on & Accuracy\\
\toprule
Depth alone & Depth & Depth  & 40.19\% \\
RGB alone & RGB & RGB  & 52.90\% \\
RGB ensemble & RGB & RGB  & 54.14\% \\
\midrule
Two-stream (average logits) & RGB+D & RGB+D  & 57.39\% \\
Two-stream after finetuning & RGB+D & RGB+D  & 58.73\% \\
ModDrop \cite{neverova2016moddrop}& \multirow{2}{*}{RGB+D} & \multirow{2}{*}{RGB+D}  & \multirow{2}{*}{58.93\%} \\
\addlinespace[-0.6ex]
(finetuned from Two-stream) & & & \\
\midrule
\midrule
ModDrop \cite{neverova2016moddrop}& RGB+D & RGB+blankD  & 47.86\% \\
ModDrop \cite{neverova2016moddrop}& RGB+D & RGB  & 53.73\% \\
Autoencoder & RGB+D & RGB  & 50.52\% \\
FCRN \cite{laina2016deeper} depth estimation & RGB+D & RGB & 50.23\% \\
Hallucination model [11]  & RGB+D & RGB & 55.94\% \\
Ours (naive adversarial)  & RGB+D & RGB & 50.81\% \\
\textbf{Ours (ADMD)} & RGB+D & RGB  &\textbf{ 57.52\%} \\
\bottomrule
\end{tabular}
\end{table}

It is well established that ensemble methods tend to outperform their single-model counterparts: an ensemble of two CNNs, each trained started from a different initialization, outperforms either independent model \cite{Guo2015DeepCE}. Since, in principle, the proposed ADMD strategy is the combination of an RGB model trained using a standard supervised approach and \textit{another} adversarially trained RGB model, we additionally compare our approach to an ensemble of RGB classifiers (third line of Table \ref{table_objrec}). Interestingly, despite starting from a two relatively high single-stream performances, the fusion process of two RGB networks only marginally increases the final accuracy (RGB1$\rightarrow$53.19\%, RGB2$\rightarrow$52.60\% $\Rightarrow$ Ensemble$\rightarrow$54.14\%).

As noticed for the task of action recognition, we found that fine-tuning the fused streams does not always bring significant improvements, as opposed  to \cite{garcia2018modality}, were the architecture features cross-stream multiplier connections, which need to be trained in an further step. Fine-tuning with the strategy proposed by Neverova \textit{et al.} \cite{neverova2016moddrop} looks slightly more effective, since ModDrop introduces a light dropout at the input layers, both on the images and on the whole modalities. The resulting model is tested in both the original setup proposed in \cite{neverova2016moddrop}, namely by blanking out the depth stream, and by simply using RGB predictions. The latter scheme slightly improves the performance of the RGB stream, possibly thanks to dropout. However, although the model shows more robustness to missing depth at test time, it clearly fails to extract any monocular depth cue.

Another interesting comparison we perform is the following: we train a cross-modal autoencoder with an L2 loss in order to reconstruct depth maps from RGB. The encoder-decoder architecture consists in the very same RGB ResNet-50 for the encoder, and in 5 stacked deconvolutional blocks intertwined with batch-norm layers for the decoder. At test time, when depth is not available, we provide RGB frames to the autoencoder, which reconstructs the missing modality to feed the corresponding branch of the two-stream architecture. The performance of this setup is quite poor. We observe that the autoencoder easily overfits the training set, generating high quality depth maps for the training set, while it performs very poorly for the test set. Similarly, we reconstruct depth by means of FCRN \cite{laina2016deeper}, a state-of-the-art depth estimator trained on the entire NYUD dataset. Again, performance is quite poor, since depth estimated by FCRN misses many fine details needed for object classification. This suggest that, for the recognition task, \textit{hallucinating task specific features is more effective than estimating depth}. 

This claim is again confirmed with the result for the Hallucination model proposed in \cite{garcia2018modality}, adapted in this case for object recognition (Table 5, 3rd to last row). This method outperforms both the RGB stream and the RGB ensemble, confirming the value of hallucinating depth. It also outperforms the other baselines that use RGB only at test time (3rd section of Table 5). In particular, it performs considerably better than FCRN depth estimation, which indicates again that depth feature hallucination is more effective than predicting depth maps at pixel level. More importantly, we can directly compare it with ADMD proposed in this paper (55.94\% vs 57.52\%), concluding that, similarly to action recognition experiments, the adversarial approach performs better.

Eventually, we tested our adversarial scheme in two different setups: i) the naive setup where the discriminator $D$ is assigned the binary task only, and ii) the ADMD setup, where the discriminator is also assigned the classification task. 
While the former performs as the autoencoder, the latter is able to fully recover the accuracy of the Two-stream model, being only slightly below that of the fine-tuned model.

\subsection{Inference with noisy depth}

\begin{table*}[h!t]
	\caption{Accuracy values for the two-stream model trained on RGB and depth, and tested with RGB and noisy depth data.}\label{tab:noise}
	\centering
	\setlength{\tabcolsep}{0em}
    \begin{tabular}{cccccccc}
		\toprule
		\multicolumn{8}{c}{NTU RGB+D action dataset - ADMD performance is 81.50\%. } \\
		&\includegraphics[width=.13\linewidth]{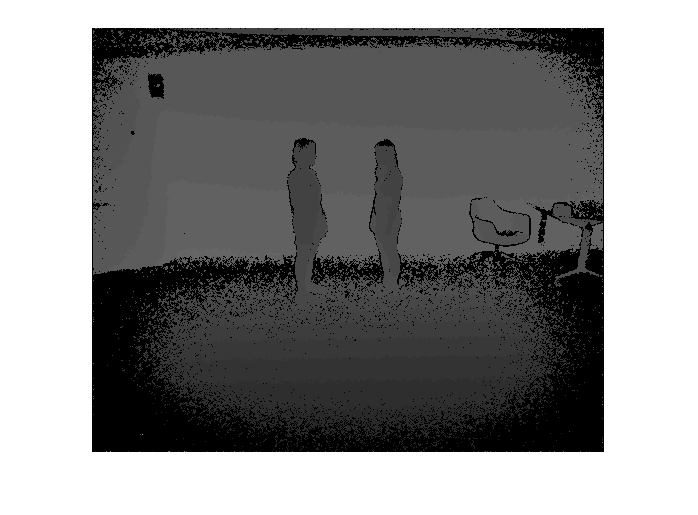}
		&\includegraphics[width=.13\linewidth]{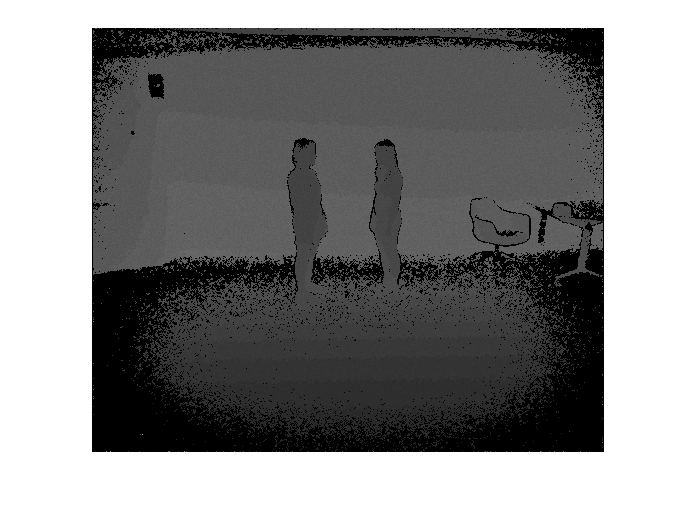}
		&\includegraphics[width=.13\linewidth]{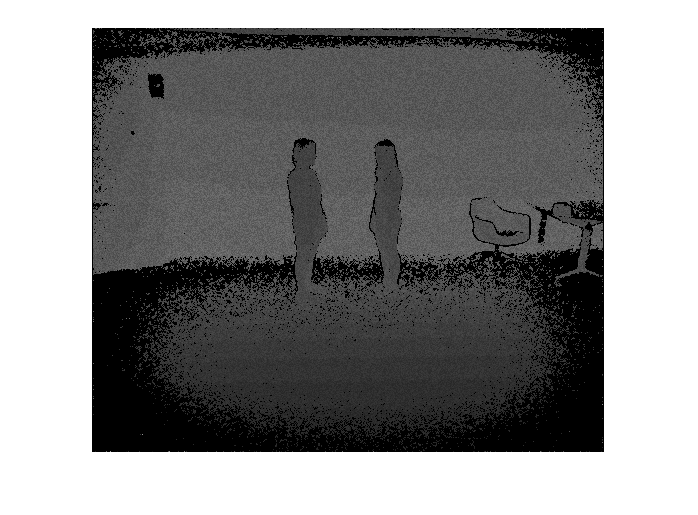}
		&\includegraphics[width=.13\linewidth]{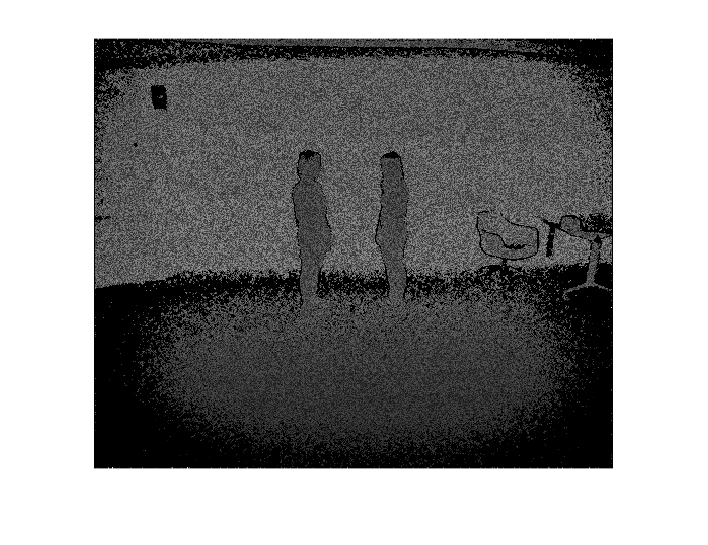}
		&\includegraphics[width=.13\linewidth]{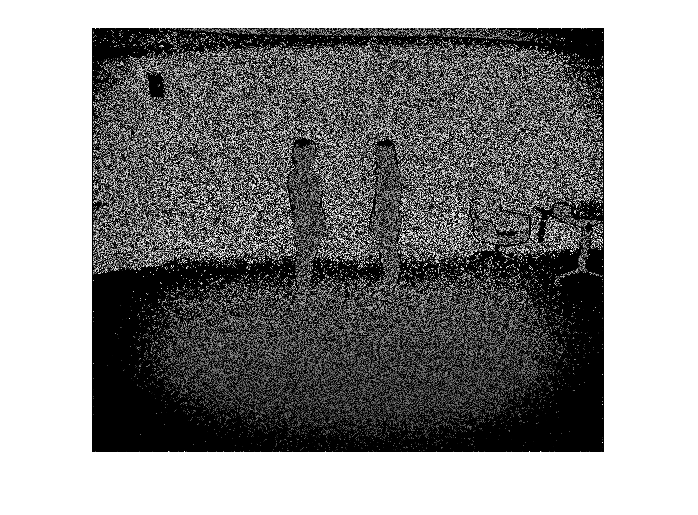}
		&\includegraphics[width=.13\linewidth]{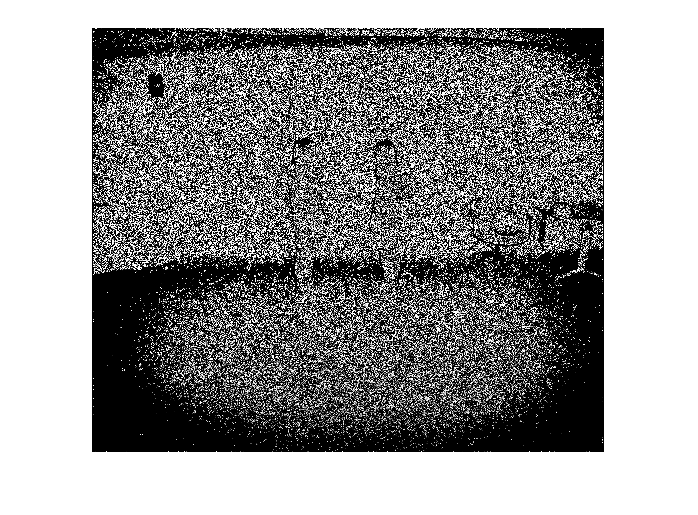}
		&\includegraphics[width=.13\linewidth]{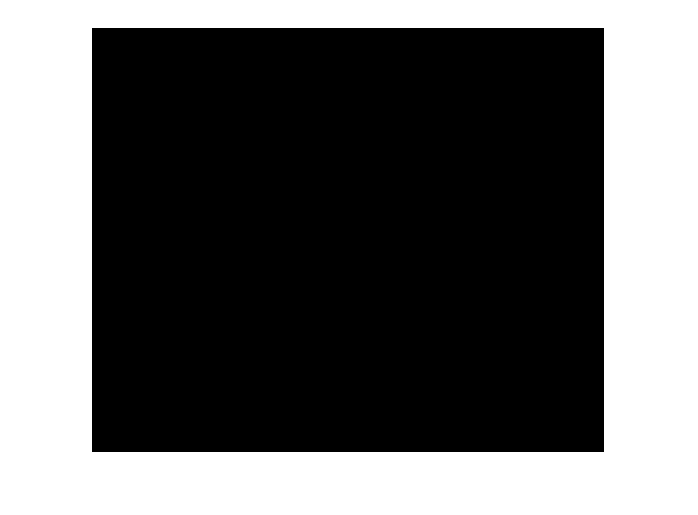}\\ 
		\addlinespace[-0.2cm]
		$\sigma^2$ & \textit{no noise} & $10^{-3}$ & $10^{-2}$& $10^{-1}$& $10^{0}$& $10^{1}$& \textit{void}\\
		Two-stream & 85.49\% & 85.52\% & 82.05\% & 68.99\% & 2.16\% & 3.35\% & 8.55\% \\
		\bottomrule
	\end{tabular} 
	\setlength{\tabcolsep}{0.2em}
	\begin{tabular}{cccccccc}
	    \vspace*{0.5ex} \\
		\toprule
		\multicolumn{8}{c}{NYUD object dataset - ADMD performance is 57.52\%. } \\
		&\includegraphics[width=.121\linewidth]{figures/NYUD/15depth.png}
		&\includegraphics[width=.121\linewidth]{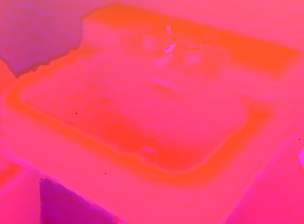}
		&\includegraphics[width=.121\linewidth]{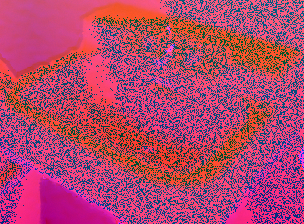}
		&\includegraphics[width=.121\linewidth]{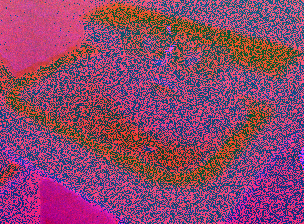}
		&\includegraphics[width=.121\linewidth]{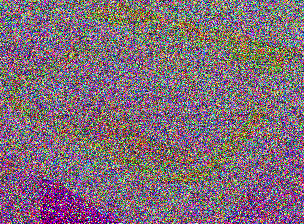}
		&\includegraphics[width=.121\linewidth]{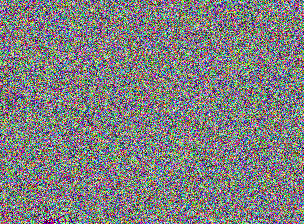}
		&\includegraphics[width=.121\linewidth]{figures/noise/void.png}\\ 
		$\sigma^2$ & \textit{no noise} & $10^{-3}$ & $10^{-2}$& $10^{-1}$& $10^{0}$& $10^{1}$& \textit{void}\\ 
		Two-stream & 58.73\% &	58.68\%	& 58.23\%	& 57.18\%	& 48.27\%	& 28.40\% & 47.44\% \\
		ModDrop \cite{neverova2016moddrop} & 58.93\% & 58.89\%	& 58.56\%	& 57.49\%	& 48.90\% &	25.95\% & 47.86\% \\
		\bottomrule
	\end{tabular} 
\end{table*}

\begin{figure}[!hb]
\centering
\includegraphics[width=\linewidth]{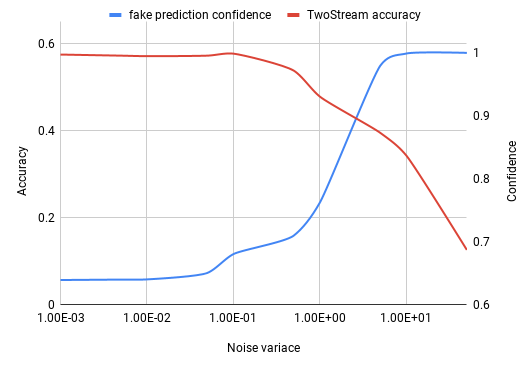}
\caption{Discriminator confidence at predicting 'fake' label as a function of noise in the depth frames. The more corrupted the frame, the more confident $D$, and the lower the accuracy of the Two-stream model (NYUD dataset).}
\label{fig:noise_D}
\end{figure}

In real test scenarios, it is often the case that we can only access \textit{noisy}  depth data. In this section, we address two questions: i) how much such noisy data can degrade the performance of a multimodal setup?  ii) At which level of noise does it become favorable to hallucinate the depth modality with respect to using the teacher model (Two-stream) with noisy depth data?

The depth sensor used in the NTU dataset (Kinect), is an IR emitter coupled with an IR camera, and has very complex noise characterization comprising at least 6 different sources \cite{kinect_noise}. 
It is beyond the scope of this work to investigate noise models affecting the depth channel, so, for our analysis we choose the most influencing one, i.e., multiplicative speckle noise. Hence, we inject Gaussian noise in the depth images $I$ in order to simulate speckle noise: $I=I * n, \, n \sim \mathcal{N}(1,\sigma) $. Table \ref{tab:noise} shows how performances of our Two-stream network degrade when depth is corrupted with such Gaussian noise with increasing variance (NTU cross-view protocol and NYUD). Results show that accuracy significantly decreases with respect to the one guaranteed by our hallucination model (81.50\% - row \#25) in Table \ref{table:sota1}, even with low noise variance of $\sigma^2$=$10^{-1}$. For the task of object recognition, we can see that ModDrop \cite{neverova2016moddrop} is slightly more resilient to depth corruption than the simple Two-stream, since fine-tuned with noise (dropout) in the input layer.

This experiment shows, in conclusion, that \textit{ADMD is able not only deal with a missing modality, but also with a noisy one}. In an online scenario, the discriminator $D$, trained in step 2, can give an indication on when to operatively switch from Two-stream to ADMD, that is, when to substitute the depth branch with the hallucination. When training reaches equilibrium, $D$ is maximally fooled by the features generated by $H$, and cannot distinguish them from those encoded by $E_d$. In practice, this means that the predicted probability for the fake class (last class in $\hat{y}$, eq. \ref{eq_extendedlabel}) is $p(\hat{y}=C+1) \approx .5$ on average. However, when features computed from corrupted depth start to flow inside $D$, its prediction for the fake class starts to be more and more confident. Figure \ref{fig:noise_D} plots the behavior of $D$ as noise increases, together with accuracy of the Two-stream model. There is a clear turning point in both accuracy and confidence, which can be employed in practice to decide when to switch from $E_d$ to $H$ \textit{i.e.} when to drop depth as a modality and start using monocular depth features extracted from RGB.

\subsection{Discussion}
Some interesting points arise from the analysis of our findings, which we summarize in the following.

\textbf{1. RGB and depth actually carry complementary information.} As a matter of fact, the Two-stream setup always provides a surprisingly better accuracy than the two streams alone. As additional evidence, a multimodal ensemble (\textit{i.e.} the Two-stream) performs better than a mono-modal ensemble (Table \ref{table_objrec}), despite the lower accuracy of one of its single-stream components (either depth or RGB, depending on task and dataset). 

\textbf{2. There is (monocular) depth information in RGB images.} This is evident from the fact that the hallucination stream often recovers and sometimes surpasses the accuracy of its depth-based teacher network. Besides, fusing hallucination and RGB streams always bring the benefits, as fusing RGB and Depth.

\textbf{3. Standard supervised learning has limitations in extracting information.} In fact, given the evidence that there is depth information 
to exploit 
in RGB images, minimizing cross-entropy loss is not enough to fully extract it. For that we need a student-teacher adversarial framework.  This has an interesting parallel in \textit{adversarial network compression} \cite{belagiannis2018adversarial}, where the performance of a fully supervised small network can be boosted by adversarial training against a high-capacity (and better performing) teacher net. In \cite{belagiannis2018adversarial}, it is also observed that the student can surpass the teacher in some occasions.

\textbf{4. Adversarial training alone only is not enough.} The naive discriminator trained for the binary task (real/generated) is not sufficient to force the hallucination network to produce discriminative features. The auxiliary discriminative task is necessary to extract monocular depth cues which are also discriminative for a given task (on the other hand, the auxiliary task only is not enough, as suggested by the performance of the RGB ensemble).

\textbf{5. Hallucinating task-specific depth features is more effective than estimating full depth images.} Not only estimated depth is often missing details needed for classification, but also its estimation is driven by mere reconstruction objectives. On the contrary, feature hallucination addresses a specific classification task and requires estimating low dimensional vectors instead of images.

\section{Conclusions}
\label{sec:concl}
In this work, we have introduced a novel technique to exploit additional information, in the form of depth images at training time, to improve RGB only models at test time.
This is done by adversarially training a hallucination network which learns from a teacher depth stream how to encode monocular depth features from RGB frames. The proposed approach outperforms previous ones in the privileged information scenario in the tasks of object classification and action recognition on three different datasets. Additionally, the hallucination framework is shown to be very effective in cases where depth is noisy.
Code is available at \url{https://github.com/pmorerio/admd}


%



\ifCLASSOPTIONcompsoc
  \section*{Acknowledgments}
\else
  \section*{Acknowledgment}
\fi

The authors would like to thank Riccardo Volpi for useful discussion on adversarial training and GANs.

\ifCLASSOPTIONcaptionsoff
  \newpage
\fi



%



{
\bibliographystyle{IEEEtran}
\bibliography{main.bib}
}

%



\begin{IEEEbiography}[{\includegraphics[width=1in,height=1.25in,clip,keepaspectratio]{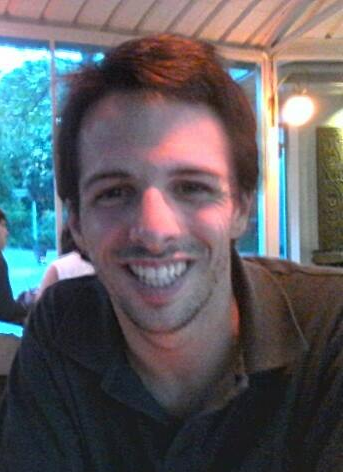}}]{Nuno C. Garcia} is a PhD fellow at the Pattern Analysis and Computer Vision department (PAVIS) in Istituto Italiano di Tecnologia (IIT) and at the Universit\`a degli studi di Genova. He was a data analytics consultant at Deloitte Portugal and a data engineer at Miniclip. He received his M. Sc. degree in Computer Engineering from the University of Beira Interior (Portugal) in 2015 (magna cum laude). His research interests are computer vision and machine learning.
\end{IEEEbiography}
\begin{IEEEbiography}[{\includegraphics[width=1in,height=1.25in,clip]{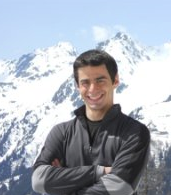}}]{Pietro Morerio} Pietro Morerio received his M. Sc. in Theoretical Physics from the University of Milan (Italy) in 2010 (summa cum laude). He was Research Fellow at the University of Genoa (Italy) from 2011 to 2012, working in Video Analysis for Interactive Cognitive Environments. He pursued a PhD degree in Computational Intelligence at the same institution in 2016. Currently he is a Postdoctoral Researcher at Istituto Italiano di Tecnologia (IIT), Pattern Analysis and Computer Vision (PAVIS) department, his research including machine learning, deep learning and computer vision, with a particular focus on multimodal learning.
\end{IEEEbiography}
\begin{IEEEbiography}[{\includegraphics[width=1in,height=1.25in,clip]{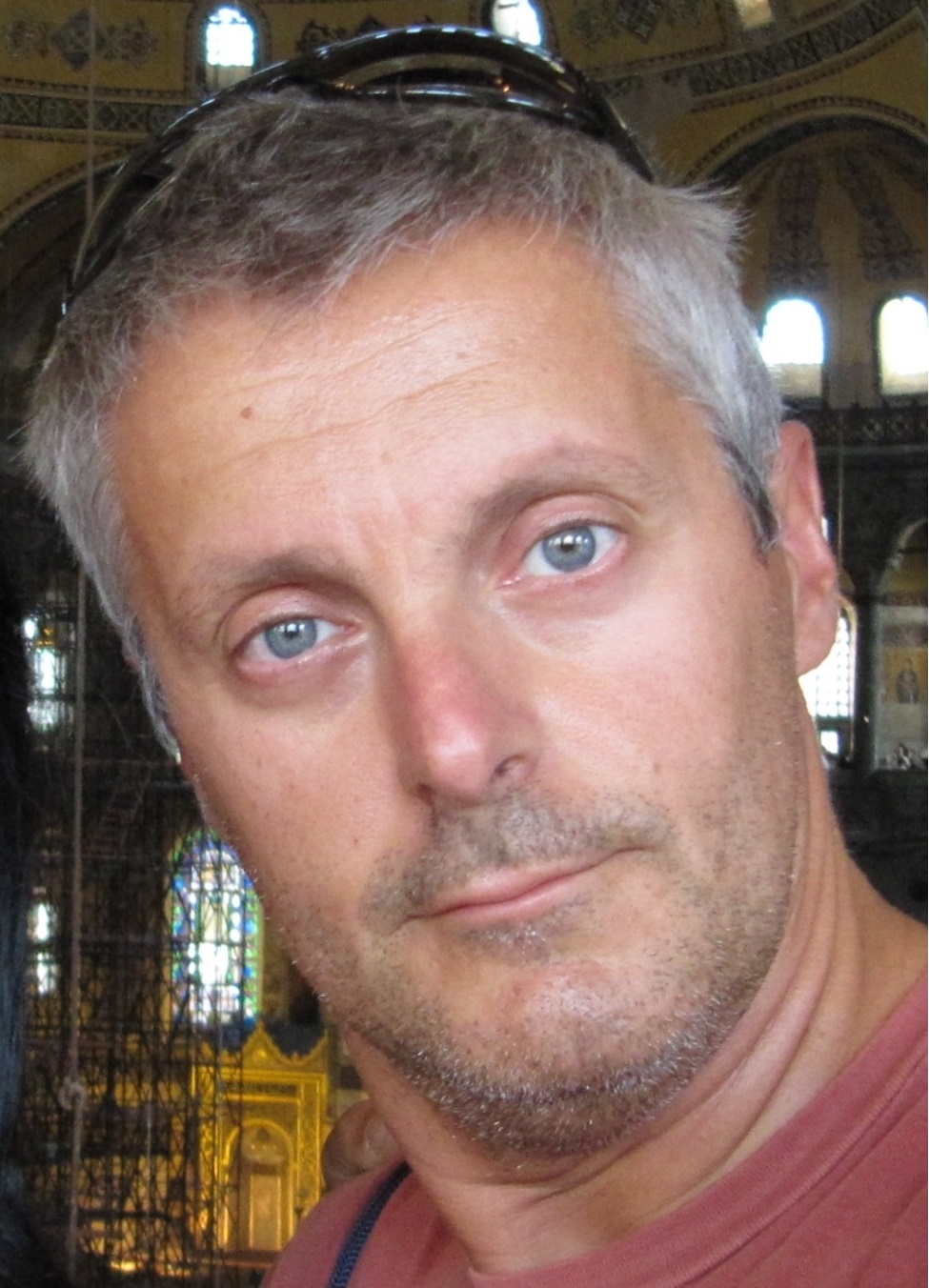}}]{Vittorio
 Murino} received the Laurea degree in electronic engineering
and the Ph.D. degree in electronic engineering and computer science from the
University of Genova, Italy, in 1989 and 1993, respectively.
He is a full professor at the University of Verona, Italy since 2000. Since 1998, he
has been with the University of Verona where he held the Chair of the
Department of Computer Science of this University from 2001 to 2007.
He is currently with the Istituto Italiano di Tecnologia, as director of Pattern
Analysis and Computer Vision (PAVIS) department, involved in computer vision,
machine learning, and image analysis activities. He has co-authored over 400
papers published in refereed journals and international conferences. His current
research interests include computer vision, pattern recognition, and machine
learning, more specifically, statistical and probabilistic techniques for image and
video processing, with applications on (human) 	behavior analysis and related
applications such as video surveillance, biomedical imaging, and bioinformatics.
He is also an Associate Editor of \textit{Computer Vision and Image Understanding}, \textit{Machine Vision \& Applications}, and \textit{Pattern Analysis and Applications} journals. He is also an IAPR Fellow since 2006.
\end{IEEEbiography}


\vfill


\end{document}